\documentclass{article}

\usepackage{arxiv}

\usepackage[utf8]{inputenc} 
\usepackage[T1]{fontenc}    
\usepackage{hyperref}       
\usepackage{url}            
\usepackage{booktabs}       
\usepackage{amsfonts}       
\usepackage{nicefrac}       
\usepackage{microtype}      
\usepackage{lipsum}
\usepackage{graphicx}
\graphicspath{ {./images/} }

\usepackage{mathtools}
\usepackage{amsfonts}
\usepackage{amsmath}
\usepackage{amssymb}
\usepackage[overload]{empheq}

\usepackage{cite}
\usepackage[authoryear]{natbib}

\newcommand{\EE}{\mathbb{E}}

\newcommand{\enstq}[2]{\left(#1~\middle|\middle|~#2\right)} 

\title{Actor loss of soft actor critic explained}

\author{
 Thibault Lahire 
}

\begin{document}
\maketitle
\begin{abstract}
This technical report is devoted to explaining how the actor loss of soft actor critic is obtained, as well as the associated gradient estimate. It gives the necessary mathematical background to derive all the presented equations, from the theoretical actor loss to the one implemented in practice. This necessitates a comparison of the reparameterization trick used in soft actor critic with the nabla log trick, which leads to open questions regarding the most efficient method to use. 
\end{abstract}


\section{Introduction}
The actor loss and the associated gradient estimate of soft actor critic \citep[SAC]{haarnoja2018soft} are not obvious to derive. The goal of this technical report is to highlight the mathematical foundations that allows deriving all equations of SAC actor (from Eq. 10 to Eq. 13). This technical report is written for all reinforcement learning \citep[RL]{sutton2018reinforcement} practitioners interested in SAC and is structured as follows. After detailing our notations in a quick set-up, we add derivations to better understand how one obtains the equation of the policy loss used in practice (Eq. 12) from the equation with the KL divergence (Eq. 10) with the help of the reparameterization trick. Then, the gradient estimate of Eq. 13 is derived, and finally compared with the gradient estimate obtained with the nabla log trick, which leads to open questions regarding the most efficient method to use.  

\section{Set-up and notations}
SAC is an off-policy actor critic algorithm, with the specificity of using regularization (hence the adjective soft). Regularization is a field of interest in RL research since efficient agents that uses it have been introduced recently \citep{vieillard2020leverage}, \citep{vieillard2020munchausen}. SAC is also an algorithm of interest when it comes to prioritizing the replay buffer \citep{wang2019boosting}, \citep{lahire2021large}, following what \citet{schaul2015prioritized} have initiated with Prioritized Experience Replay.

In SAC paper, states and actions are denoted with the subscript $t$ to indicate the timestep. Since we do not need it in this article, we will drop the subscript. $Q_\theta$ is the critic network, parameterized by $\theta$. It takes as input a state $s$ and an action $a$, and outputs a scalar. $\pi_\phi$ is the actor network, parameterized by $\phi$. It takes as input a state $s$ and outputs a probability distribution over the action space. The best action to take from state $s$ is then sampled from the probability distribution $\pi_\phi(\cdot|s)$.

The policy loss is defined as the KL divergence between the probability distribution over the action space proposed by the actor network and the probability distribution induced by the exponentiated current Q function. Since the exponentiated current Q function is not a probability distribution (the integral over the action space does not sum to 1), a normalization factor $Z_\theta$ is used.

\section{From KL divergence to generic policy loss}
The goal of this KL divergence is to measure the gap between the distribution of the actor network and the distribution induced by the exponentiated current Q function. Since the high values of the latter indicates the areas of the action space where the cumulative expected sum of rewards is approximated to be high, minimizing the KL divergence means getting an efficient actor network associated to actions yielding highly rewarded trajectories. Using the definition of the KL divergence, the following can be derived.

\begin{align*}
J_{\pi}(\phi) &=  \EE_{s \sim \mathcal{D}} \left[ D_{KL}\enstq{\pi_{\phi}(\cdot|s)}{\frac{\exp Q_{\theta}(s, \cdot)}{Z_{\theta}(s)}}\right] \\
&= \EE_{s \sim \mathcal{D}} \left[ \int_{a} \pi_{\phi}(a|s) \log \frac{\pi_{\phi}(a|s)}{\exp(Q_{\theta}(s, a))/Z_{\theta}(s)} da \right] \\
&= \EE_{s \sim \mathcal{D}} \left[ \int_{a} (\pi_{\phi}(a|s) \log \pi_{\phi}(a|s) - \pi_{\phi}(a|s) Q_{\theta}(s, a)) da \right] + C\\
&= \EE_{s \sim \mathcal{D}} \left[ \EE_{a \sim \pi_{\phi(\cdot|s)}}[\log \pi_{\phi}(a|s) - Q_{\theta}(s, a)] \right] + C \qquad \qquad \qquad \qquad \qquad \qquad (\star)
\end{align*}

where $C$ is a constant with respect to $\phi$. The last equation is very close to Eq. 12 in SAC paper, but still different. To obtain Eq. 12, the reparameterization trick must be used. 

\section{Reparameterization trick}

We restrict our explanations to the gaussian case, since our interest lies in it. To sample the random variable $a \sim \mathcal{N}(\mu, \sigma)$, one can simply sample $\epsilon \sim \mathcal{N}(0,1)$ and then apply the transformation $a = \mu + \sigma \epsilon$. The reparameterization trick has its roots in this equation. 

The SAC’s authors have made the choice of using a gaussian distribution for $\pi_\phi$. Concretely, the actor network outputs a mean $\mu_\phi(s)$ and a standard deviation $\sigma_\phi(s)$ when the state $s$ is given as input. $\mu_\phi(s)$ and $\sigma_\phi(s)$ have the same dimension, and are both vectors. The variance-covariance matrix of the gaussian considered is diagonal, and $\sigma_\phi(s)$ is its diagonal. In the following, we use $\mathcal{N}(\mu_\phi(s), \sigma_\phi(s))$ to refer to the probability law according to which actions are drawn, even though it is a slight abuse of notation since $\sigma_\phi(s)$ is not a matrix.

The reparameterization trick proposes to sample the action by sampling $\epsilon \sim \mathcal{N}(0,1)$ and then applying the linear transformation $a = \mu_\phi(s) + \epsilon \sigma_\phi(s)$. The right hand-side of this equation is written $f_\phi(\epsilon;s)$ in SAC paper (Eq. 11). By sampling according to $\mathcal{N}(0,1)$ instead of $\mathcal{N}(\mu_\phi(s), \sigma_\phi(s))$, the expectation of Eq. ($\star$) above changes, and we obtain Eq. 12 of SAC paper:

$$\EE_{s \sim \mathcal{D}} \left[ \EE_{a \sim \pi_{\phi(\cdot|s)}}[\log \pi_{\phi}(a|s) - Q_{\theta}(s, a)] \right] = \EE_{s \sim \mathcal{D}} \left[ \EE_{\epsilon \sim \mathcal{N}(0,1)}[\log \pi_{\phi}(f_{\phi}(\epsilon;s)|s) - Q_{\theta}(s, f_{\phi}(\epsilon;s))] \right]. $$

\section{Derivation of the gradient estimate}

Thanks to the reparameterization trick, the gradient of the policy loss is straightforward: \\$\nabla_\phi J_\pi(\phi) = \mathbb{E}_{s \sim \mathcal{D}} \left[ \mathbb{E}_{\epsilon \sim \mathcal{N}(0,1)} \left[ \nabla_\phi \log \pi_\phi(f_\phi(\epsilon;s)|s) - \nabla_\phi  Q_\theta(s, f_\phi(\epsilon;s)) \right] \right]$.

The chain rule applied on the second term yields: $\nabla_\phi  Q_\theta(s, f_\phi(\epsilon;s)) = \nabla_a Q_\theta(s, a) \nabla_\phi f_\phi(\epsilon; s).$

For the first term, however, the gradient with respect to $\phi$ of a more complex function must be taken. This function depends directly on $\phi$ (because $\phi$ parameterizes $\pi$), but also indirectly with $f_\phi$. Let us re-write $f$ as $f(\phi, \epsilon, s)$ and $\log \pi_\phi(f_\phi(\epsilon;s)|s)$ as $g(\phi, f(\phi, \epsilon, s), s)$ to highlight the dependence on $\phi$. Let us assume, for simplicity, that $\phi$ is a scalar. The extension to the vector case is straightforward. 

To obtain the gradient, let us write the second order Taylor development of function $g$. It yields:

\begin{align*}
    g(\phi + \Delta \phi, f(\phi + \Delta \phi, \epsilon, s), s) =& \ g(\phi, f(\phi, \epsilon, s), s) + \frac{\partial g}{\partial \phi}(\phi, f(\phi, \epsilon, s), s) \left( \phi + \Delta \phi - \phi \right) \\
    &+ \frac{\partial g}{\partial f}(\phi, f(\phi, \epsilon, s), s) \left( f(\phi + \Delta \phi, \epsilon, s) - f(\phi, \epsilon, s) \right) + o(\Delta \phi^2).
\end{align*}

The rate of change with respect to variable $\phi$ is:

\begin{align*}
    &\frac{g(\phi + \Delta \phi, f(\phi + \Delta \phi, \epsilon, s), s) - g(\phi, f(\phi, \epsilon, s), s)}{\Delta \phi} = \frac{\partial g}{\partial \phi}(\phi, f(\phi, \epsilon, s), s) \\
    &\qquad\qquad\qquad\qquad\qquad\qquad\qquad\qquad+ \frac{\partial g}{\partial f}(\phi, f(\phi, \epsilon, s), s) \frac{ f(\phi + \Delta \phi, \epsilon, s) - f(\phi, \epsilon, s) }{\Delta \phi} + o(\Delta \phi).
\end{align*}

By definition, the derivative of $g$ with respect to $\phi$ is the limit as this rate of change goes to zero. It yields: 

$$\frac{d g}{d \phi}(\phi, f(\phi, \epsilon, s), s) = \frac{\partial g}{\partial \phi}(\phi, f(\phi, \epsilon, s), s) + \frac{\partial g}{\partial f}(\phi, f(\phi, \epsilon, s), s) \frac{d f}{d \phi}(\phi, \epsilon, s).$$ 

Note the key difference between the $\frac{d \quad}{d \quad}$ operator and the $\frac{\partial \quad}{\partial \quad}$ operator. The first one corresponds to the total derivative, which is the one we are interested in for the gradient computation, whereas the second one corresponds to a partial derivative. $\frac{\partial g}{\partial \phi}(\phi, f(\phi, \epsilon, s), s)$ means that the variables $f(\phi, \epsilon, s)$ and $s$ are considered constant, whereas $\phi$ is the variable along which the derivative applies. The SAC’s authors decided to use $a$ instead of $f$ to insist on the non-variability: $\frac{\partial g}{\partial \phi}(\phi, a, s)$. When switching to the vector case for $\phi$, it yields:

$$\nabla_\phi J_\pi(\phi) = \mathbb{E}_{s \sim \mathcal{D}} \left[ \mathbb{E}_{\epsilon \sim \mathcal{N}(0,1)} \left[ \nabla_\phi \log \pi_\phi(a|s)  \right] \right] + \mathbb{E}_{s \sim \mathcal{D}} \left[ \mathbb{E}_{\epsilon \sim \mathcal{N}(0,1)} \left[ \left( \nabla_a \log \pi_\phi(a|s) - \nabla_a Q_\phi(s,a) \right) \nabla_\phi f_\phi(\epsilon;s)  \right] \right].$$\\

To approximate the first expectation over $s$, a state $s$ is randomly drawn from the replay buffer. To approximate the second expectation over $\mathcal{N}(0,1)$, a sample $\epsilon$ is drawn from this distribution, from which the action $a$ is derived. This yields Eq. 13:

$$\hat{\nabla}_\phi J_\pi(\phi) =  \nabla_\phi \log \pi_\phi(a|s) +  \left( \nabla_a \log \pi_\phi(a|s) - \nabla_a Q_\phi(s,a) \right) \nabla_\phi f_\phi(\epsilon;s). $$\\

Note that this analytical gradient is useless in practice with common deep learning frameworks such as pytorch \citep{paszke2017automatic} or tensorflow \citep{abadi2016tensorflow}. Indeed, these libraries are such that writing the policy loss is enough: the gradient is automatically computed when the backward pass is performed. 

\section{Discussion and open questions}

Note that using a gaussian policy is a choice. In this section, we weight the pros and cons of such a choice. The probability distribution induced by the exponentiated current Q function is not necessarily gaussian and can be more complex, for example multimodal. Hence, the gaussian policy used in SAC might poorly approximate the target probability distribution. As an illustration, suppose a one dimensional action space, upon which the normalized exponentiated current Q function is a mixture of two gaussian distributions of standard deviation 1 and centered in -2 and 2. Let $h$ be this distribution. Constraining the policy to be a single gaussian is a very poor proxy for this multimodal distribution. Let $\pi_\phi$ be this gaussian of standard deviation 1 and mean $\phi$, and let’s find the optimal scalar $\phi$ minimizing the KL divergence $KL(h||\pi_\phi)$. The minimum of this KL divergence is $\phi = 0$, which means that most actions will be around 0, whereas they should be around -2 or 2. 

This issue can be alleviated by using a mixture of gaussian distributions as a policy. However, the reparameterization trick cannot be used in this case. Instead, the nabla log trick is typically used, details are provided below. From $J_\pi(\phi) = \mathbb{E}_{s \sim \mathcal{D}} \left[ \mathbb{E}_{a \sim \pi_\phi(\cdot|s)} \left[ \log \pi_\phi(a|s) - Q_\theta(s, a) \right] \right]$, it yields:
\begin{align*}
    \nabla_\phi J_\pi(\phi) &= \mathbb{E}_{s \sim \mathcal{D}} \left[ \int_a \pi_\phi(a|s) \nabla_\phi \log \pi_\phi(a|s) da + \int_a \log \pi_\phi(a|s) \nabla_\phi \pi_\phi(a|s) da - \int_a Q_\theta(s, a) \nabla_\phi \pi_\phi(a|s) da \right] \\
    &= \mathbb{E}_{s \sim \mathcal{D}} \left[ \mathbb{E}_{a \sim \pi_\phi(\cdot|s)} \left[ \nabla_\phi \log \pi_\phi(a|s) \right] \right] + \mathbb{E}_{s \sim \mathcal{D}} \left[ \int_a (\log \pi_\phi(a|s)  -  Q_\theta(s, a)) \nabla_\phi \pi_\phi(a|s) da \right] \\
    &= \mathbb{E}_{s \sim \mathcal{D}} \left[ \mathbb{E}_{a \sim \pi_\phi(\cdot|s)} \left[ \nabla_\phi \log \pi_\phi(a|s) \right] \right] + \mathbb{E}_{s \sim \mathcal{D}} \left[ \int_a (\log \pi_\phi(a|s)  -  Q_\theta(s, a)) \pi_\phi(a|s) \nabla_\phi \log \pi_\phi(a|s) da \right] \\
    &= \mathbb{E}_{s \sim \mathcal{D}} \left[ \mathbb{E}_{a \sim \pi_\phi(\cdot|s)} \left[ \nabla_\phi \log \pi_\phi(a|s) \right] \right] + \mathbb{E}_{s \sim \mathcal{D}} \left[ \mathbb{E}_{a \sim \pi_\phi(\cdot|s)} \left[ (\log \pi_\phi(a|s)  -  Q_\theta(s, a))  \nabla_\phi \log \pi_\phi(a|s) \right] \right]. 
\end{align*}

\textbf{How many gaussian distributions for the mixture?} The SAC’s authors have tested several mixture of gaussian distributions in the first version of their paper (which can be found on arXiv), with different number of gaussian distributions. The results are reported on Fig. 5.b, for a test realized on the Mujoco \citep{todorov2012mujoco} HalfCheetah-v1 task. It appears that using a mixture of gaussian distributions does not bring improvement upon a single gaussian. However, it is possible that the HalfCheetah-v1 task is well suited for a gaussian policy. To verify that a single gaussian is indeed enough, more tasks should be tested.

\textbf{Nabla log trick or reparameterization trick?} This is the open question on which we conclude this technical report. The reparameterization trick is often said to yield lower variance of the gradient estimate than the nabla log trick, but this claim is not supported by any theoretical proof in general. As explained above, the nabla log trick allows using a mixture of gaussian distributions as policy, which improve the quality of approximation of the exponentiated current Q function. However, setting the number of gaussian distributions of the mixture remains a difficult task.

\bibliographystyle{unsrtnat}  
\bibliography{arxiv}  


\end{document}